\documentclass[a4paper, 10pt, conference]{ieeeconf}

\newif\ifoagmfinalcopy

\oagmfinalcopytrue

\IEEEoverridecommandlockouts

\usepackage{graphics} 
\usepackage{epsfig} 
\usepackage{mathptmx} 
\usepackage{times} 
\usepackage{amsmath} 
\usepackage{amssymb}  

\usepackage{todonotes}

\usepackage{lineno}
\usepackage{tikzpagenodes}
\usepackage{background}
\usepackage{url}

\ifoagmfinalcopy
\backgroundsetup{color=white}
\else

\linenumbers

\newcommand{\MyOAGMConfidentialLogo}{
\begin{tikzpicture}[remember picture,overlay]
\node[align=center,text=blue] at ([yshift=1em]current page text area.north) {\Large \#\#\# ARW/OAGM 2021 SUBMISSION: CONFIDENTIAL REVIEW COPY \#\#\#};
\end{tikzpicture}
}

\SetBgContents{\MyOAGMConfidentialLogo}
\SetBgPosition{current page.north west}
\SetBgOpacity{0.5}
\SetBgAngle{0.0}
\SetBgScale{1.0}

\fi

\title{\LARGE \bf
Explaining YOLO: Leveraging Grad-CAM to Explain Object Detections
}

\ifoagmfinalcopy
\author{Armin Kirchknopf$^{1}$, Djordje Slijepčević$^{1}$, Ilkay Wunderlich$^{2}$, Michael Breiter$^{2}$, \\Johannes Traxler$^{2}$, and Matthias Zeppelzauer$^{1}$ 
\thanks{*This research was funded by the Austrian Research Promotion Agency (FFG) project 876468 ``SAiEX'',  \protect\url{https://bit.ly/saiex}}
\thanks{$^{1}$Institute of Creative Media Technologies, St. Pölten University of Applied Sciences, Austria, {\tt\small firstname.lastname@fhstp.ac.at}.}
\thanks{$^{2}$EYYES GmbH, Austria, {\tt\small firstname.lastname@eyyes.com}.}
}%
\else
\author{Anon, Ymous}
\fi

\begin{document}

\maketitle

\begin{abstract}

We investigate the problem of explainability for visual object detectors. Specifically, we demonstrate on the example of the YOLO object detector how to integrate Grad-CAM into the model architecture and analyze the results. We show how to compute attribution-based explanations for individual detections and find that the normalization of the results has a great impact on their interpretation.

\end{abstract}

\section{INTRODUCTION}

Today's complex computer vision models require mechanisms that explain their behavior. This has fueled intensive research in eXplainable Artificial Intelligence (XAI)~\cite{adadi2018peeking}. Most work on XAI in the visual domain focuses on explaining visual classifiers, i.e., their representations learned and/or their decisions. Currently, there is  a lack of XAI approaches for visual object detectors, because their special architectures impede the application of XAI methods.

In this paper, we investigate the problem of XAI for visual object detectors on the example of the YOLO detector~\cite{redmon2018yolov3}. We integrate Grad-CAM~\cite{Selvaraju_2019} into the model to generate explanations for individual object detections, i.e., bounding boxes. We compute attention maps at detection level to assess which information leads to a certain decision. For this purpose, we focus on both scores estimated by the YOLO detector, namely \textit{objectness} and \textit{class probability}, to obtain a more comprehensive explanation. We critically analyze the results and propose different normalization strategies to make the attention maps of different object detections within an input image or across different images comparable. We analyze results obtained for true and false  detections and compare different normalization variants for result presentation. 

There is a large corpus of related work both on object detection~\cite{liu2020deep} and on XAI~\cite{adadi2018peeking}. Surprisingly, the combination of both fields has hardly been investigated. Rare exceptions are the work of Tsunakawa et al. \cite{tsunakawa2019contrastive}, who proposed an extension of a propagation-based XAI method (Layer-wise Relevance Propagation, LRP) for Single Shot MultiBox Detectors and Petsiuk et al.~\cite{petsiuk2021black}, who proposed a post-hoc model-agnostic XAI method for object detectors based on randomized input sampling. The lack of literature may be a consequence of the highly specific architectures of object detectors that impedes the integration of XAI methods. Object detectors require the explanation of localization and classification aspects and provide multiple scores that influence the likelihood of a detection. This makes the direct application of many, especially self-learned explainability approaches~\cite{chen2018learning}, difficult. More promising candidates are post-hoc XAI approaches. 

A popular example  is LIME~\cite{ribeiro2016should}, which can be adapted easily to explain the final output of a detector. To explain internal scores, the direct application is, however, not possible. Additionally, the iterative probing approach of LIME makes it slow. A faster and more flexible approach is Grad-CAM, which propagates back the activation of a certain neuron (an output neuron or some intermediate neuron) to the last feature map of the underlying convolutional filter stack and uses it to weight its activations. The weighted activations in the last feature map can be directly up-scaled and overlaid with the input image to obtain an attribution-based explanation in terms of the high-level features learned by the convolutional filter stack. Note that this is more meaningful than back-propagating along the gradients completely through the network until the input pixels (guided Grad-CAM), as individual pixels lack semantic meaning.

\section{METHOD}

Our detection model is based on Tiny YOLO v3~\cite{redmon2018yolov3} architecture with optimizations for inference on re-configurable hardware~\cite{wunderlich2019adaptations} and contains two detection heads to account for objects with different scales. The last convolutional layer of each head stores multiple scores for each potential bounding box: (i) \textit{objectness}, which provides the likelihood for observing an object in general and (ii) a vector of \textit{class probabilities} for all target classes. For head 1, this layer has a  size of 1x1x512x30 and for head 2 1x1x256x30. Specific neurons in these  layers represent the input to Grad-CAM for the generation of explanations. After these layers the YOLO architectures applies a non-maximum suppression (NMS) and a decision threshold filters out the most likely detections.

\begin{figure} [h!]
    \centering
    \includegraphics[width=0.8\linewidth]{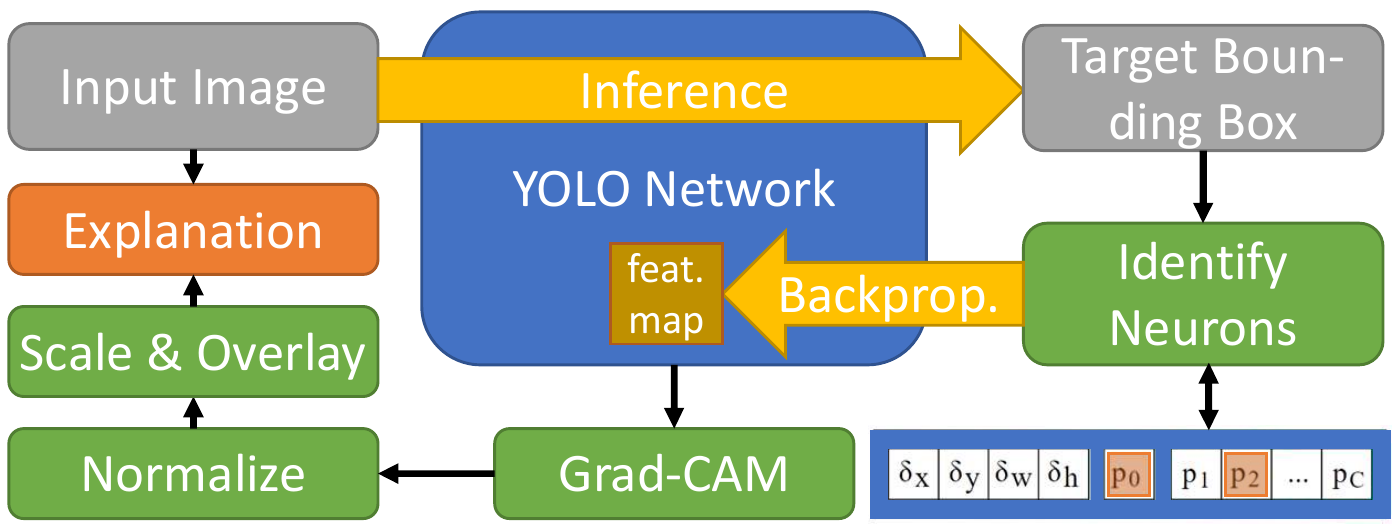}
    \caption{Proposed explainability approach. Here $p_0$ represents the objectness neuron and $p_2$ the target class neuron in the last layer of the respective detection head.}
    \label{fig:approach}
\end{figure}

\begin{figure*} [h!]
    \centering
    \includegraphics[width=0.75\linewidth]{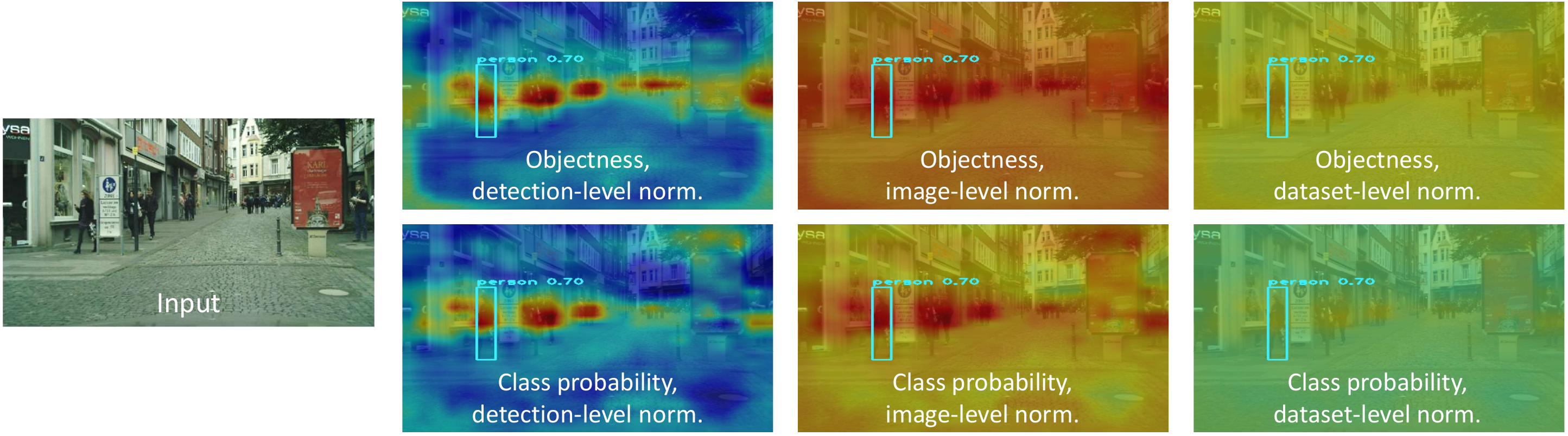}
    \caption{Explanations for a true positive detection for objectness and class probability and three different normalization variants.}
    \label{fig:resultTP}
\end{figure*}

Grad-CAM was originally proposed for conventional CNN architectures to explain decisions in terms of abstract features learned in the last convolutional layer. Considering that YOLO is based on a convolutional filter stack, Grad-CAM is applicable, however, not without certain modifications. For a given detection, we first identify the neurons in the last convolutional layer of the respective head corresponding to the class probability and objectness of the investigated bounding box by reversing the NMS process. These neurons represent the starting points to calculate gradients towards the neurons of the underlying convolutional layer (i.e., the top-level feature map of the convolutional stack). We follow a two-step approach to obtain explanations for both scores. The gradients are first used to weight the activation map of the underlying convolutional layer. The weighted activation map is then averaged over all channels of the layer and upscaled (i.e., interpolation) and mapped (i.e., color coding) to the input image (416px x 416px), see Figure~\ref{fig:approach}. The upscaled activation pattern highlights sections in the input image that have a strong relation to the class or objectness of the investigated bounding box. Note that due to the architecture of YOLO the result of Grad-CAM are activations at the global image level, i.e., they are not limited to the observed bounding box, e.g., as shown in Figure~\ref{fig:resultTP}.

Grad-CAM activations are by default min-max normalized to improve visibility. This leads to incomparable activations patterns between different object detections in the \textit{same} image and across \textit{different} images. To account for this, we propose three different normalization levels: detection-level (default), image-level (joint normalization of all explanations in an image), and dataset-level (joint normalization of all explanations across a set of images).

\section{EXPERIMENTS AND RESULTS}

\paragraph{Model training}

The network was trained on data of front collision and rearview cameras from both public datasets including COCO, KITTI, BDD, and OpenImages as well as non-public data from the company EYYES GmbH (\url{www.eyyes.com}).
The network was trained to detect five  classes, i.e., person, cycle, car, truck, and train.

\paragraph{Experimental Setup} 

Our evaluation scenario originates from autonomous driving. For the evaluation, we use a subset of the Cityscapes (\url{www.cityscapes-dataset.com/}) dataset (which was not used for training). It consists of 3470 images showing urban street scenes from 21 cities and containing annotations of 30 classes at pixel-level. We use a subset of the above mentioned five classes. For the different normalization strategies we use min-max normalization of the Grad-CAM activations at different levels.

\paragraph{Results} Results are shown as differently normalized heatmaps overlaid on the input image for the objectness and the probability of the detected class. Figure~\ref{fig:resultTP} shows a correct detection of a person. The objectness shows a different activation pattern than the class probability. While class probability provides strong activations mostly on persons (including the detected one), objectness activates on all regions where the network sees potential objects. Results for detection-level normalization are most distinct, which can, however, lead to wrong conclusions, especially, when the explanation shall be compared with other detections. It actually depends on the question investigated which type of normalization is best suited. For example by normalization at dataset-level the activation strength of the detected person becomes directly comparable to all explanations  in all other images and thus direct comparisons become possible which cannot be performed at detection level. This can help to develop a deeper understanding of the detector's behavior.  

\begin{figure} [h!]
    \centering
    \includegraphics[width=1\linewidth]{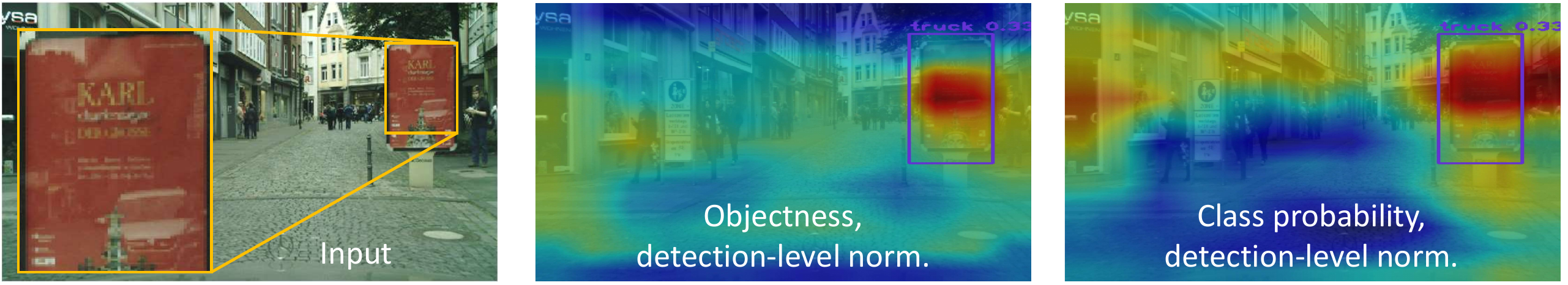}
    \caption{Explanations for a false positive detection.}
    \vspace{-5pt}
    \label{fig:resultFP}
\end{figure}

Figure~\ref{fig:resultFP} shows a falsely detected truck on the same input image. The white rectangular shaped text on the red poster seems to mislead the detector into seeing a truck. Both objectness and class probability strongly activate at detection-level raising the impression that the detector fails with high confidence. This is actually not true, which can be seen via normalization at dataset-level (not shown) where both activations are strongly attenuated, showing that the detector is actually not sure about the detection.

\vspace{-5pt}
\section{CONCLUSION AND FUTURE WORK}

We have investigated explainability for object detection by integrating Grad-CAM into YOLO. We can visualize its internal decision scores and thereby help to explain object detections. Our results show that normalization is essential to make different explanations comparable, e.g., across different images. Our approach is efficient: generating one explanation takes approx. half a second. In future, we aim to use these explanations to identify potential false detections at run-time.

{\small
\bibliographystyle{IEEEtranS}
\bibliography{IEEEexample}
}

\end{document}